\documentclass[conference]{IEEEtran}
\usepackage{times}

\usepackage{jj}
\usepackage[numbers]{natbib}
\usepackage{multicol}
\usepackage[bookmarks=true]{hyperref}
\usepackage{cleveref}
\usepackage{balance}

\begin{document}

\title{Realizing Robotic Swimming with Unified Fluid-Robot Multiphysics}


%
\author{\authorblockN{Jeong Hun Lee\textsuperscript{1}\authorrefmark{1},
Junzhe Hu\textsuperscript{1}\authorrefmark{1},
Sofia Kwok\textsuperscript{1}, 
Carmel Majidi\textsuperscript{1} and
Zachary Manchester\textsuperscript{1}\textsuperscript{2}}
\authorblockA{\textsuperscript{1}Carnegie Mellon University, \textsuperscript{2}Massachusetts Institute of Technology,
\authorrefmark{1}Equal Contribution}
\authorblockA{Project Page: \url{https://unified-fluid-robot-multiphysics.github.io/}}
}
\makeatletter

\let\old@maketitle\@maketitle

\renewcommand{\@maketitle}{
  \old@maketitle
  \begin{center}
    \includegraphics[width=0.99\linewidth]{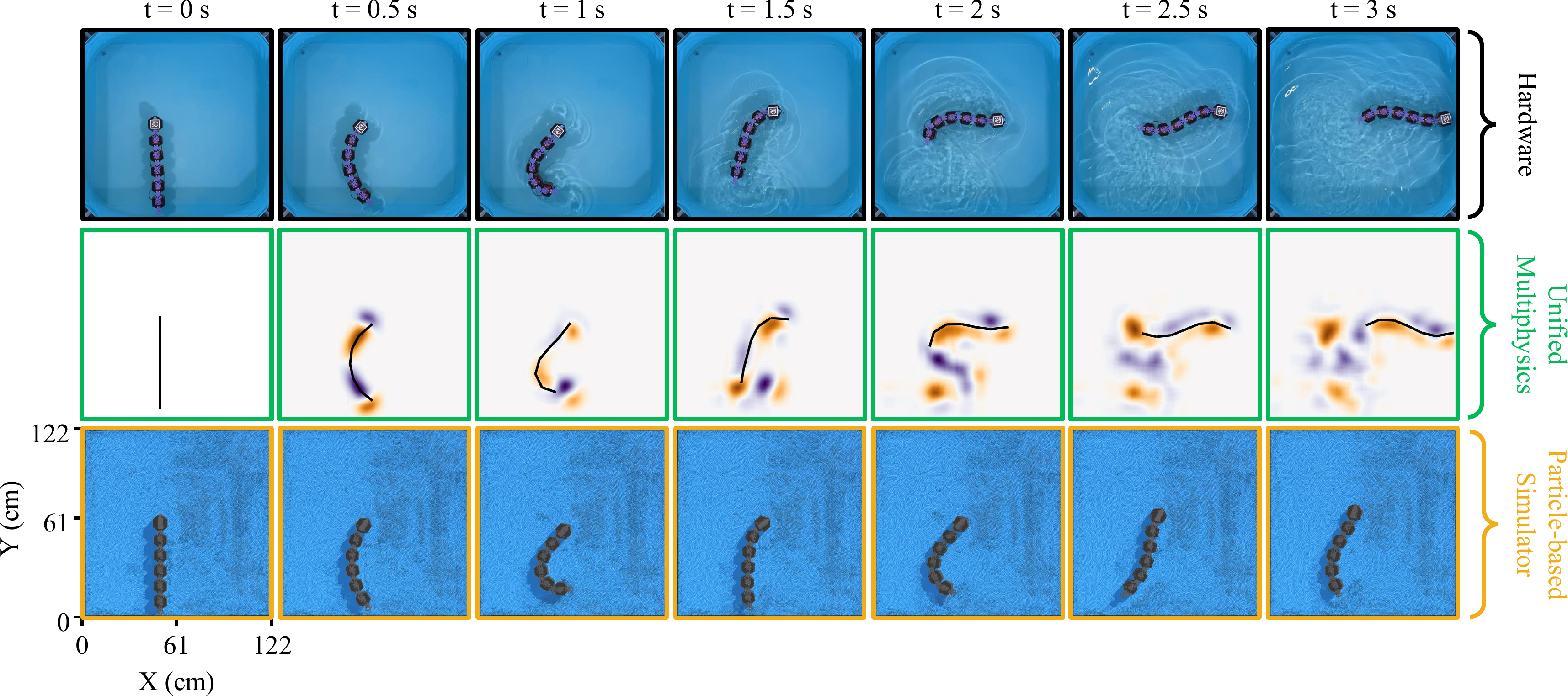}
    \captionof{figure}{A highly dynamic fish-inspired C-start escape maneuver executed by a multilink eel-robot. The top row (black) shows execution of the maneuver on the robot, while the middle row (\color{jj_green}green\color{black}) is our simulation method and the bottom row (\color{jj_orange}orange\color{black}) is a baseline implemented in Genesis~\cite{genesisauthorsGenesisUniversalGenerative2024a}, an open-source robotics simulator. The C-start gait was optimized using the gradients from our differentiable simulator and was successfully executed on real-world hardware. Meanwhile, the smoothed-particle hydrodynamics baseline from Genesis fails to capture the correct dynamics---rolling out the same control trajectory achieves little lateral movement and does not complete the intended $90^\circ$ turn.
    }
    \label{fig:c_start_trajectory_visual}
    \vspace{\baselineskip}
  \end{center}
  \vspace{-0.12in}
}


\let\oldmaketitle\maketitle

\renewcommand{\maketitle}{

  \oldmaketitle

  \addtocounter{figure}{-1}

}

\maketitle

\makeatother

\begin{abstract}
Matching the swimming efficiency and agility of fish has remained an elusive goal in underwater robotics. Such locomotion capabilities rely on complex vortex interactions between the robot's body and the surrounding fluid. However, simulating these dynamics, which are governed by coupled ordinary and partial differential equations, is significantly more difficult than the multi-body dynamics of classical rigid robotic systems. We present a differentiable framework for simulating strongly coupled fluid-robot multiphysics as a unified optimization problem. The coupled manipulator and incompressible Navier-Stokes equations are derived together from a single Lagrangian using the principle of least action. We employ discrete variational mechanics to derive a stable, well-conditioned, and physically accurate scheme for jointly simulating articulated bodies and the surrounding fluid. We leverage the implicit function theorem to compute derivatives of the fully coupled dynamics. Using this simulator and its gradients, we realize undulating swimming gaits and optimize a highly dynamic C-start escape maneuver for a bioinspired eel robot. We validate both gaits on physical hardware, demonstrating successful sim-to-real transfer. Simulation code, hardware data, and schematics for the eel robot can be found here: \url{https://unified-fluid-robot-multiphysics.github.io/}
\end{abstract}\href{}{}

\IEEEpeerreviewmaketitle

\section{Introduction}

In recent years, there has been considerable interest in designing robot control policies in multiphysics settings such as deformable-object manipulation~\cite{xuDextAIRityDeformableManipulation2022, shiRobocookLonghorizonElastoplastic2023, yooRoPotterRoboticPottery2024, xingStabilizingReinforcementLearning2024}, fluid transport~\cite{lagrassaTaskOrientedActiveLearning2024, ichnowskiGompfitGraspoptimizedMotion2022}, and locomotion~\cite{yangMultiphysicsDynamicModel2019, costaDesignCarangiformSwimming2020, oconnellNeuralFlyEnablesRapid2022}. However, real-world data collection and training in such complex environments is challenging. As a result, multiphysics simulation platforms~\cite{macklinWarpHighperformancePython2022, genesisauthorsGenesisUniversalGenerative2024a, comsol} have seen growing interest and ongoing development, with many robotics applications still lacking viable physics engines.

\begin{figure}[t]
    \centering
    \includegraphics[width=0.49\textwidth]{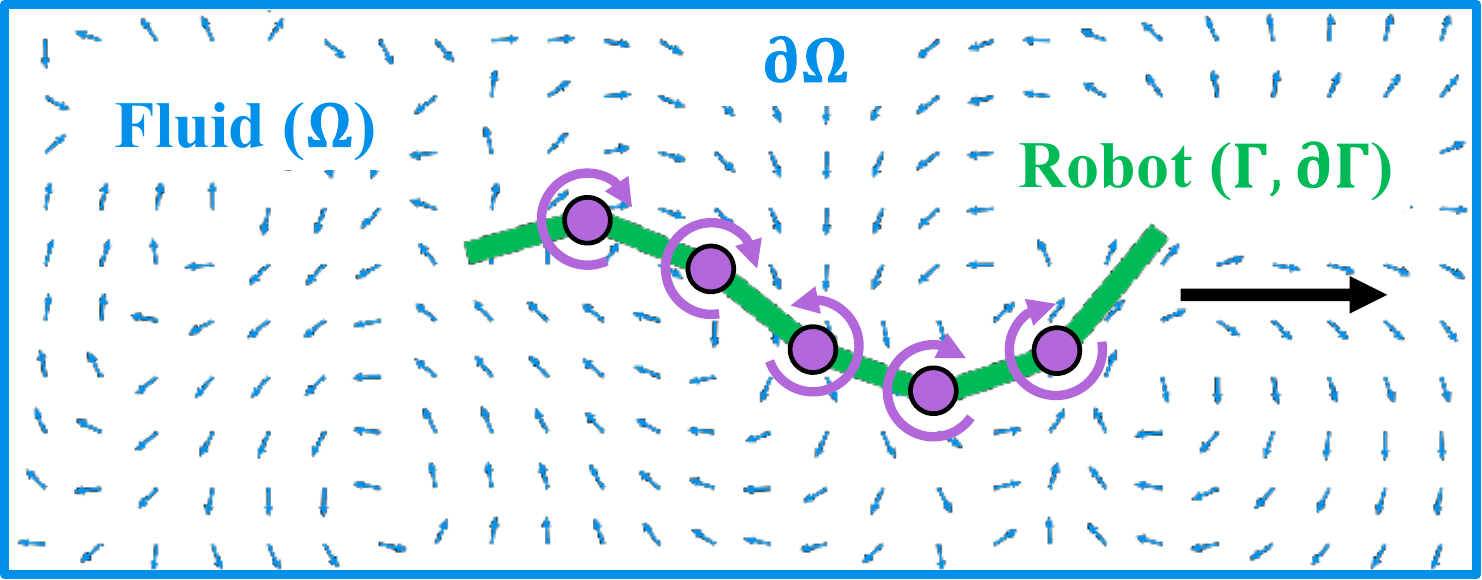}
    \caption{An overview of the multiphysics problem we address, in which a controlled, rigid-body robotic system must interact with the surrounding incompressible Newtonian fluid (e.g., water) to swim. We denote the fluid domain as $\Omega$ and its boundary $\partial \Omega$. $\Gamma$ and $\partial \Gamma$ correspond to the robot geometry and its boundary.}
    \label{fig:problem_setup}
    \vspace{-\baselineskip}
\end{figure}

Bio-inspired locomotion for underwater vehicles~\cite{maertensOptimalUndulatorySwimming2017, gaoIndependentCaudalFin2018, novatiSynchronisationLearningTwo2017, katzschmannExplorationUnderwaterLife2018a, milesDontBeJelly2019a, linLearningAgileSwimming2025} has garnered significant interest as a means to improve efficiency and maneuverability via intricate shedding of vortices~\cite{triantafyllouEfficientSwimmingMachine1995b, triantafyllouHydrodynamicsFishlikeSwimming2000a, bealPassivePropulsionVortex2006a}. Although various bio-inspired hardware designs have been introduced~\cite{andersonOscillatingFoilsHigh1998a, katzschmannExplorationUnderwaterLife2018a, andersonManeuveringStabilityPerformance2002a, christiansonCephalopodinspiredRobotCapable2020a}, relatively little work has been done to make these vehicles operate autonomously with the same performance as their biological counterparts~\cite{andersonManeuveringStabilityPerformance2002a, white2021tunabot}. We attribute this to the difficulty of modeling and simulating such systems, where both whole-body robot dynamics and complex fluid-structure interaction (FSI) are critical (Fig.~\ref{fig:problem_setup}).

\begin{figure}[t]
    \centering
    \includegraphics[width=0.49\textwidth]{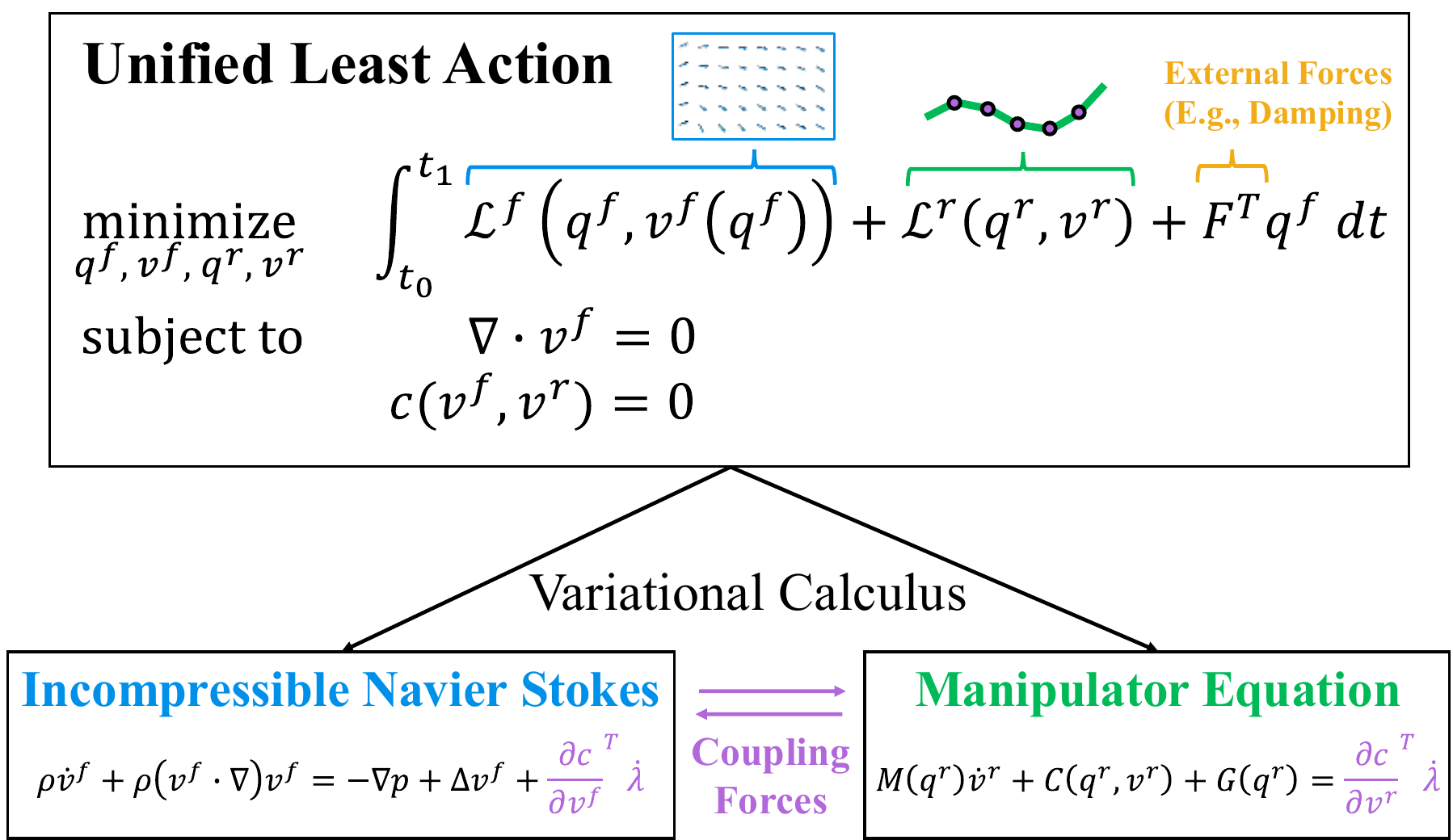}
    \caption{An overview of our multiphysics model, in which the unified Lagrangian governing the coupled fluid-robot dynamics is formulated using the principle of least action. Rather than the individual differential equations, the combined action is discretized directly to achieve consistent strongly coupled simulation.}
    \label{fig:least_action_diagram}
    \vspace{-\baselineskip}
\end{figure}

We propose a unified framework for jointly deriving and simulating differentiable fluid-robot multiphysics to realize robotic swimming. Specifically, we formulate the combined Lagrangian using the principle of least action to model the multiphysics as a single continuous-time optimization problem, from which we derive the coupled incompressible Navier-Stokes and manipulator equations (Fig.~\ref{fig:least_action_diagram}). We show that the multiphysics coupling stems from a constraint that enforces the no-slip boundary condition at the fluid-robot interface. We then employ discrete variational mechanics~\cite{marsdenDiscreteMechanicsVariational2001} to discretize the unified action directly, which results in an implicit time-integration scheme. We also extend the immersed boundary method from the computational fluid dynamics (CFD) literature~\cite{tairaImmersedBoundaryMethod2007a, peskinImmersedBoundaryMethod2002a} to derive an integral ``weak-form'' version of the no-slip constraint that is amenable to multibody robotic systems. The implicit function theorem is then used to compute simulation derivatives for downstream learning and optimization tasks. Using our unified multiphysics, we realize bioinspired swimming on an eel robot, which we validate in real-world hardware experiments. We also benchmark against a smoothed-particle hydrodynamics (SPH) environment implemented in Genesis~\cite{genesisauthorsGenesisUniversalGenerative2024a}, a state-of-the-art robotics multiphysics simulator. We investigate both steady undulatory swimming and the highly dynamic C-start escape maneuver, which we optimize to achieve a 90-degree turn. In summary, our contributions are:

\begin{itemize}

  \item A unified derivation of strongly coupled fluid-robot multiphysics via the principle of least action.

  \item A new weak formulation of the no-slip boundary constraint at the fluid-robot interface.
    
  \item A differentiable fluid-robot multiphysics simulator.
  
  \item Realization of accurate, dynamic robotic swimming on a bioinspired eel robot with validation against real-world hardware.
  
\end{itemize}

The remainder of this paper is organized as follows: In \Cref{sec:RELATED WORKS}, we provide a literature review on bioinspired swimming, fluid-robot interaction simulators, and variational mechanics. In \Cref{sec:BACKGROUND}, we review background material on the least-action principle, variational integrators, and the immersed-boundary method. \Cref{sec:METHODOLOGY} then describes our approach for posing and simulating strongly coupled fluid-robot multiphysics. In \Cref{sec:RESULTS}, we provide simulation and hardware swimming results for an eel-like swimming robot. Finally, \Cref{sec:CONCLUSION} summarizes our conclusions and discusses limitations and directions for future work.

\section{RELATED WORKS} \label{sec:RELATED WORKS}

\subsection{Bioinspired Robotic Swimming}

For decades, biological swimmers have captivated researchers with their remarkable swimming efficiency~\cite{grayStudiesAnimalLocomotion1936, lighthillLargeamplitudeElongatedbodyTheory1971, fish2006passive} and high maneuverability~\cite{webbFastStartPerformanceBody1978, domeniciEscapeResponsesFish2019}. While early work focused on biomechanics~\cite{grayStudiesAnimalLocomotion1936, lighthillLargeamplitudeElongatedbodyTheory1971, webbFastStartPerformanceBody1978}, later studies attributed this performance to the detection and exploitation of vortices~\cite{triantafyllouHydrodynamicsFishlikeSwimming2000a, triantafyllouEfficientSwimmingMachine1995b}. This has driven considerable interest in the robotics community toward developing bioinspired hardware for empirical studies, which include robotic fish~\cite{andersonManeuveringStabilityPerformance2002a, white2021tunabot, katzschmannExplorationUnderwaterLife2018a}, oscillating foils~\cite{andersonOscillatingFoilsHigh1998a}, and lateral-line pressure-sensor arrays~\cite{venturelli2012hydrodynamic, yangArtificialLateralLine2010}. However, relatively little work has enabled these swimmers to operate with the robustness achieved by legged robots~\cite{mikiLearningRobustPerceptive2022a, kumarRMARapidMotor2021, zhangWholeBodyModelPredictiveControl2025}---a gap we attribute to the lack of simulators capable of handling both complex fluid physics and rigid-body robot dynamics~\cite{liuFishGymHighPerformancePhysicsbased2022b, angelidisSmoothedParticleHydrodynamics2025, genesisauthorsGenesisUniversalGenerative2024a}.

\subsection{Fluid Simulation for Robotics}

Multiphysics simulators for robotics~\cite{macklinWarpHighperformancePython2022, genesisauthorsGenesisUniversalGenerative2024a} have garnered recent interest for simulating diverse scenes beyond classical rigid-body dynamics, such as rigid-soft-body interactions for deformable-object manipulation~\cite{qiaoDifferentiableSimulationSoft2021} and liquid pouring~\cite{xianFluidLabDifferentiableEnvironment2023}. These simulators generally employ methods originating from the graphics and particle-based CFD community~\cite{macklinUnifiedParticlePhysics2014, huMovingLeastSquares2018}, leveraging their parallelizability and visual-rendering capabilities. However, due to their Lagrangian (i.e., particle-based) representations, these simulators are tailored towards fluid-transport tasks (e.g., liquid pouring) encountered in robotic manipulation. Compared to their Eulerian (grid-based) counterparts, these fluid simulators are currently unable to simulate complex boundary conditions (e.g., free stream) and become intractable when the fluid must be modeled as a continuum. In addition, SPH-based methods can struggle with maintaining certain constraints such as incompressibility and conservation of volume, which are better posed in Eulerian-based methods~\cite{nairVolumeConservationIssues2015, suchdeVolumeMassConservation2025, guermondOverviewProjectionMethods2006}.

\subsection{Fluid-Structure Interaction (FSI)}

Modeling fluid-structure interaction has long been of interest in the CFD community~\cite{palaciosStanfordUniversityUnstructured2014a, navaFastAquaticSwimmer2022, liuFishGymHighPerformancePhysicsbased2022b, leeAquariumFullyDifferentiable2023}, where methods are typically tailored to single-body systems. Recently, ~\citet{angelidisSmoothedParticleHydrodynamics2025} proposed an SPH-based platform for bioinspired, multibody robotic swimmers. However, the authors note that smoothed-particle hydrodynamics introduces additional damping and parameter sensitivity, causing unstable simulation and sim-to-real gap. Meanwhile, Eulerian methods capable of handling articulated multibody systems commonly found in robotics have yet to be fully realized~\cite{todorovMuJoCoPhysicsEngine2012b}. Previous works have simplified the fluid dynamics to potential~\cite{jiaoLearningSwimPotential2021} or Stokes flow~\cite{duFunctionalOptimizationFluidic2020, groverGeometricMotionPlanning2018, groverMotionPlanningDesign2019} for low-Reynolds-number regimes. Nava et al. proposed a physics-informed, neural-network model of FSI for the optimization of soft robotic swimmers~\cite{navaFastAquaticSwimmer2022}; however, accuracy is generally poor when generalizing to new shapes and flow conditions. The industry-standard COMSOL multiphysics solver~\cite{comsol} provides strongly-coupled, Eulerian-based FSI via the Arbitrary Lagrangian-Eulerian method, but requires computationally expensive remeshing that struggles with large displacements~\cite{mittal2023origin}. As a result, \citet{leeAquariumFullyDifferentiable2023} proposed Aquarium, a differentiable incompressible-Navier-Stokes-based simulator for robotics. However, Aquarium currently does not jointly solve the fluid and robot dynamics and relies on a separate dynamics solver to obtain robot states (i.e., ''weakly'' coupled), which can result in unstable simulation~\cite{heil2008solvers}. It additionally relies on the classical immersed-boundary method to represent robot geometry, which is tailored toward single-body systems and suffers from singularities when bodies move close together. Therefore, we develop a unified multiphysics formulation to achieve tightly coupled, generalizable simulation of multibody robots in higher-Reynolds-number fluid environments.

\subsection{Variational Mechanics for Simulation}

Over the past two decades, variational integrators have been developed for accurate simulations of various physical systems~\cite{lewOverviewVariationalIntegrators1970, marsdenDiscreteMechanicsVariational2001, howellDojoDifferentiableSimulator2022}. Specifically, variational integrators discretize the least-action principle from which the governing equations of motion are derived. The result is stable simulation with good constraint satisfaction and momentum and energy-conservation properties. Although originally for conservative systems, variational integrators have been extended to damped systems and continuum mechanics~\cite{lewOverviewVariationalIntegrators1970}, with connections made to widely used implicit Runge-Kutta schemes~\cite{marsdenDiscreteMechanicsVariational2001}. Recent efforts have extended variational integrators to robotics in the context of multibody simulation with rigid-body contact seen in legged locomotion and manipulation~\cite{howellDojoDifferentiableSimulator2022}. However, variational integrators remain an unconventional method for physics simulation, which we partly attribute to their configuration-only representation. Therefore, we extend the variational integrator to a fluid-robot multiphysics setting with full robot-state representations found in standard robotics simulators~\cite{todorovMuJoCoPhysicsEngine2012b}.

In the context of fluid dynamics, variational principles have long been studied~\cite{arnoldTopologicalMethodsHydrodynamics2009, seligerVariationalPrinciplesContinuum1968} to pose fluid physics as continuous-time optimization problems. This optimization-based treatment has inspired several practices in CFD, including the treatment of boundary conditions as constraints with associated dual variables~\cite{papanastasiouNewOutflowBoundary1992, allmarasLagrangeMultiplierImplementation2005, tairaImmersedBoundaryMethod2007a}.
We aim to extend these variational principles to the discrete setting with robot dynamics in mind, resulting in naturally coupled multiphysics simulation.

\section{BACKGROUND} \label{sec:BACKGROUND}

This section provides a brief overview of variational calculus, variational integrators, and the immersed-boundary method for fluid-structure interaction. We refer the reader to the comprehensive, existing literature for more details~\cite{verziccoImmersedBoundaryMethods2023, marsdenDiscreteMechanicsVariational2001, arnoldTopologicalMethodsHydrodynamics2009, seligerVariationalPrinciplesContinuum1968}.

\subsection{Physics as Optimization}

Various differential equations for modeling physics can be derived from the principle of least action via variational calculus. Specifically, the principle of least action represents the dynamics as a continuous-time optimization problem:
\begin{mini}|l|
  {q, v}{\int_{t_0}^{t_f} T\big(v(t)\big) - U\big(q(t)\big) \, dt}{}{} \label{eq:simple_la_principle}
  \addConstraint{v = \dot{q},}
\end{mini}
where $q(t) \in \R^{n}$ is the configuration of the system (e.g., fluid-particle position and robot pose) as a function of time, $t$; $v(t) \in \R^{m}$ is the velocity; $T(v) \in \R+$ is the kinetic energy of the system; and $U(q) \in \R+$ is the potential energy; and $v=\dot{q}$ represents the kinematic constraint. We denote $L(q, v)=T(v)-U(q)$ as the \emph{physics} Lagrangian, which we can use to derive an expression for the action,
\begin{equation}
    S(q, v) = \int_{t_0}^{t_f} L(q, v) + \lambda^T (\dot{q} - v) \, dt, \label{eq:generic_action}
\end{equation}
where $\lambda$ is the dual variable.
The first-order necessary (FON) conditions of the optimization problem \eqref{eq:simple_la_principle} form the Euler-Lagrange equation that governs the system dynamics:
\begin{equation}
    \frac{d}{dt} \frac{\partial L}{\partial \dot{q}} - \frac{\partial L}{\partial q} = 0, \label{eq:euler_lagrange}
\end{equation}
which is commonly rewritten as the manipulator equation in robotics:
\begin{equation}
    M(q) \ddot{q} + C(q, \dot{q})j + G(q) = 0. \label{eq:manipulator equation}
\end{equation}

We note that the least-action principle can be extended to non-conservative systems via the Lagrange-D'Alambert principle of virtual work~\cite{lewOverviewVariationalIntegrators1970}; to more complex kinematics $v=f(q, \dot{q})$ as encountered e.g. with different parametrizations of attitude~\cite{jacksonPlanningAttitude2021}; and to include constraints such as joints in manipulator arms~\cite{baraffLineartimeDynamicsUsing1996}:

\begin{mini}|l|
  {q,v}{\int_{t_0}^{t_f} T(q,v) - U(q) + F(t)^T q(t)\, dt}{}{} \label{eq:simple_la_principle}
  \addConstraint{v = f(q,\dot{q})}
  \addConstraint{c(q, v) = 0,}
\end{mini}
where $F(t) \in \R^n$ represents external forces such as damping and motor torques.

\subsection{Variational Integrators}

Robotics simulators commonly integrate \eqref{eq:manipulator equation} via a Runge-Kutta (RK) scheme~\cite{todorovMuJoCoPhysicsEngine2012b}. However, most RK schemes are known to exhibit unstable behavior or introduce artificial damping. Meanwhile, variational integrators provide implicit time-integration schemes that are known to conserve energy and momentum, making them attractive for robotics simulators.

\begin{figure}[t]
    \centering
    \includegraphics[width=0.48\textwidth]{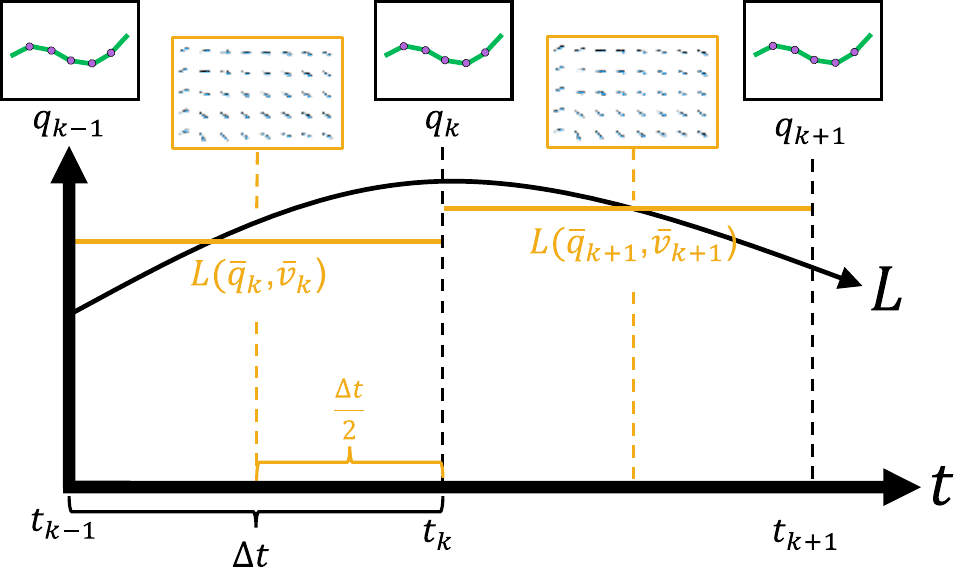}
    \caption{A midpoint variational integrator, where a quadrature rule is traditionally used to discretize the time integral of Fig.~\ref{fig:least_action_diagram} to simulate robot configurations, $q_{k}$. However, the fluid is represented as a velocity field, $\bar{v}_{k}$ that is approximated at the midpoint, causing a temporal mismatch. Therefore, a new variational integrator is formulated for the fluid-robot multiphysics.}
    \label{fig:variational_integrator}
    \vspace{-\baselineskip}
\end{figure}

First, the continuous-time action integral is discretized via numerical quadrature. As a particular example (Fig.~\ref{fig:variational_integrator}), a midpoint quadrature rule results in the following optimization problem:
\begin{mini}|l|
  {q_k}{\sum_{k=1}^{2} L_{d}(q_{k}, q_{k+1}) + \tfrac{h}{2}F(t_{k} + \tfrac{h}{2})^T (q_k + q_{k+1})}{}{} \label{eq:discrete_least_action}
  \addConstraint{c(q_{k}) = 0,}
\end{mini}
where $q_k$ is the configuration at time $t_k$; $h \in \R$ is the time step; and $L_{d}(q_{k}, q_{k+1}) \in \R+$ is the discrete Lagrangian, which is expressed as:
\begin{equation}
    L_{d}(q_{k}, q_{k+1}) = h L(\underbrace{\tfrac{q_k + q_{k+1}}{2}}_{\Bar{q}_{k+1}}, \underbrace{\tfrac{q_{k+1} - q_k}{h}}_{\Bar{v}_{k+1}}) ,\label{eq:midpoint_rule}
\end{equation}
where $\Bar{q}_{k+1}$ and $\Bar{v}_{k+1}$ are the configurations and velocities defined at the midpoint between $t_{k}$ and $t_{k+1}$.

We then obtain the discrete Euler-Lagrange equation, which is the FON condition of \eqref{eq:discrete_least_action} w.r.t. $q_k$:
\begin{multline}
    D_{2}L_{d}(q_{k-1}, q_{k}) + D_{1}L_{d}(q_{k}, q_{k+1}) \\
    + h\Bar{F}_{k} + h\underbrace{\lambda_{k}^T D_{1} c(q_k)}_{\text{constraint force}} = 0, \label{eq:discrete_euler_lagrange}
\end{multline}
\begin{equation}
    c(q_{k+1}) = 0, \label{eq:discrete_constraint}
\end{equation}
where $\lambda_{k} \in \R^{z}$ is the dual variable. Upon inspection, $\lambda_k$ provides the impulse needed to enforce the constraint $c(q_{k+1})=0$. $D_{k}$ denotes partial differentiation with respect to a function's $k$th argument, and $\Bar{F}_k$ represents the averaged external force $F(t)$ at the midpoints:
\begin{equation}
    \Bar{F}_k = \tfrac{1}{2}(F(t_k - \tfrac{h}{2}) + F(t_k + \tfrac{h}{2})).
\end{equation}

By initializing $q_{k}$ and $q_{k-1}$, \eqref{eq:discrete_euler_lagrange}-\eqref{eq:discrete_constraint} can be solved to compute $q_{k+1}$. However, this integration method only integrates over configurations, with a half-time-step delay present between $q_k$ and $\Bar{v}_k$. This introduces a temporal inconsistency due to the fluid being modeled as a velocity field, $\Bar{v}_k(q_k)$, while the robot's pose is defined by $q_k$. \citet{marsdenDiscreteMechanicsVariational2001} previously connected this variational integrator to the implicit midpoint RK scheme over the full state, $x = [q; v]$, via the discrete Legendre transform. However, doing so with constraints can be difficult and unintuitive. Therefore, we aim to provide a simpler approach for performing variational integration over the full variational state, $x = [q; \bar{v}]$.

\begin{figure}[t]
    \centering
    \includegraphics[width=0.42\textwidth]{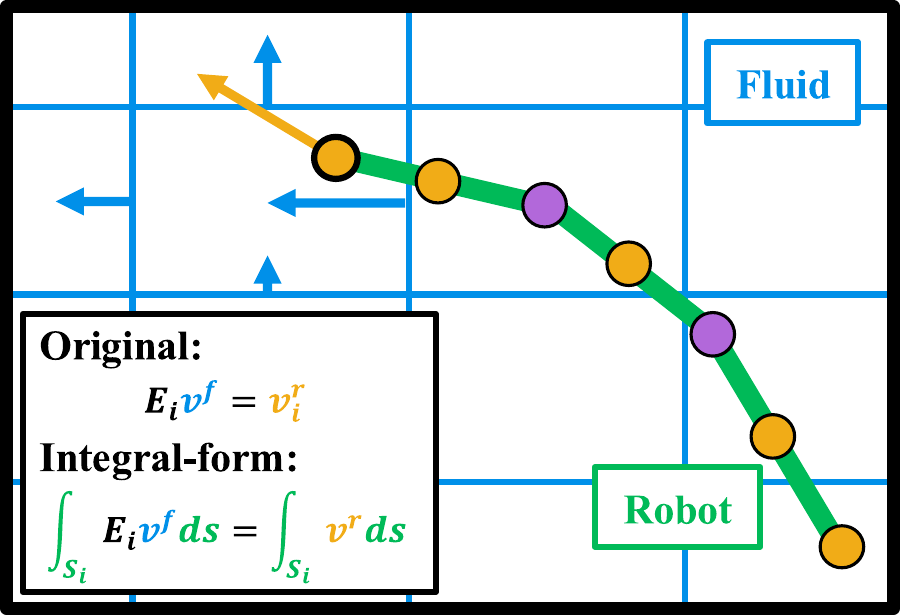}
    \vspace{0.1\baselineskip}
    \caption{The immersed boundary method, where the fluid domain and robot geometry are discretized separately into a grid and boundary-mesh. Originally, the fluid-robot coupling is realized by a convolution matrix $E$ that maps fluid-cell velocities (\color{jj_blue}blue\color{black}) to those at the robot-boundary nodes (\color{jj_orange}orange\color{black}). This can lead to fluid-penetration and singularities at duplicate nodes (\color{jj_purple}purple\color{black}) that correspond to multiple bodies. Therefore, we introduce an extension that integrates the immersed boundary method across the boundary (\color{jj_green}green\color{black}), improving both physical accuracy and compatibility with multibody representations, which are common in robotics.}
    \vspace{-\baselineskip}
    \label{fig:immersed_boundary}
\end{figure}

\subsection{The Immersed Boundary Method}

To account for the representation discrepancy between the Eulerian fluid-velocity grid and Lagrangian (free-floating) robot geometry, \citet{peskinImmersedBoundaryMethod2002a} formulated the immersed boundary method to model the no-slip boundary condition at the robot-fluid interface as shown in Fig.~\ref{fig:immersed_boundary}. Specifically, the fluid velocity is set equal to the robot velocity at its boundary, which is expressed as the following integration over the fluid-velocity field:
\begin{equation}
    \int_{\Omega} v^{f} \delta (q^f - \prescript{b}{}{q}^r) \, d\Omega = \prescript{b}{}{v}^r(v^r) \quad \forall \, \prescript{b}{}{q}^r \in \partial \Gamma, \label{eq:continuous_immersed_boundary}
\end{equation}
where $\Omega$ is the fluid domain; $\partial \Gamma$ is the robot geometry's boundary; $v^f$ is the fluid velocity at position $q^f$; $q^r, v^r$ are the solid body (i.e., robot) configuration and velocity; and $\prescript{b}{}{v}^r$ is the velocity at a point $\prescript{b}{}{q}^r$ along $\partial \Gamma$. $\delta()$ is the dirac delta function. In addition, the corresponding forcing term acting on the fluid is explicitly defined as an inverse mapping:
\begin{equation}
    F^f = \int_{\Gamma} F^r\delta (q^f - \prescript{b}{}{q}^r) \, d\Gamma \quad \forall \, q^f, \label{eq:continuous_immersed_boundary_force}
\end{equation}

In discrete form, the fluid domain and robot geometry are discretized into a separate grid and mesh respectively, allowing for decoupled movement of the robot. Meanwhile, the dirac delta is approximated by a smoothed version. This results in the discretized immersed boundary method:
\begin{align}
    c(v^f, v^r) & = Ev_{d}^f - \prescript{b}{}{v}_d^r(v^r) = 0, \label{eq:discrete_immersed_boundary} \\
    F_d^f & = E^T F_d^r, \label{eq:discrete_immersed_boundary_force}
\end{align}    
where $E$ is a convolution matrix that maps discrete fluid velocities on the grid to those of the robot-boundary nodes. However, upon inspection, this discretized constraint is only satisfied at finite nodes of the boundary mesh, allowing fluid to penetrate the boundary if the nodes are far apart. On the other hand, overlapping points can cause numerical ill-conditioning due to redundant no-slip constraints, which can occur in multibody robots as shown in Fig.~\ref{fig:immersed_boundary}. Therefore, we extend the immersed boundary method to derive an integral-form that will properly enforce no-slip in between nodes while being unaffected by overlapping ones.


\section{METHODOLOGY} \label{sec:METHODOLOGY}

\subsection{Fluid-Robot Multiphysics as Optimization}

As previously mentioned, existing simulators~\cite{macklinWarpHighperformancePython2022, genesisauthorsGenesisUniversalGenerative2024a} typically integrate over the individual differential equations before coupling the physics via a force term such as $\eqref{eq:continuous_immersed_boundary_force}$. We propose the least-action principle as an alternative, unified perspective of multiphysics from which the differential equations are derived.

We begin by proposing the least-action principle for the combined multiphysics problem between a rigid multibody robot and an incompressible, Newtonian fluid (e.g., water):
\begin{small}
    \begin{mini}|l|
      {q^r_k, q^f, v^r, v^f}{\int_{t_0}^{t_f} L^{f}(q^f, v^f) + L^r(q^r, v^r) + F(t)^Tq^f \, dt}{}{} \label{eq:combined_fluid_robot_optimization}
      \addConstraint{c_1(v^f) = \dot{q}^f - v^f(q^f) = 0 \quad \forall \, q^f \in \Omega}
      \addConstraint{c_2(v^f) = \nabla \cdot v^f = 0 \quad \forall \, q^f \in \Omega}
      \addConstraint{c_3(v^f) = v^f - v_{bc} = 0 \quad \forall \, q^{f} \in \partial\Omega}
      \addConstraint{c_4(v^r) = \dot{q}^r - v^r = 0 \quad \forall \, q^r}
      \addConstraint{c_5(q^r) = 0 \quad \forall \, q^r}
      \addConstraint{c_6(v^f, v^r) = v^f - \prescript{b}{}{}v^r = 0, \quad \forall \, q^f \in \Gamma,}
    \end{mini}
\end{small}
where $\Omega, \Gamma$ represent the fluid domain and robot geometry respectively with corresponding boundaries, $\partial\Omega, \partial\Gamma$; $q^r, q^f$ are the configurations of the robot and fluid particle respectively and $v^r, v^f$ are the corresponding velocities. $c_{1}(v^f)$ encodes the kinematics constraint for the fluid-velocity field (i.e., an array of velocity sensors). $c_{2}(v^f)$ encodes the conservation-of-mass constraint while $c_{3}(v^f)$ enforces boundary conditions along the fluid-domain boundary (e.g., walls, free-stream velocity, etc.). $c_4(q^r)$ encodes the kinematics constraint of the robot while $c_5$ enforces other robot-configuration constraints, such as joint constraints found in manipulator arms. $c_6(v^f, v^r)$ encodes the no-slip constraint, which couples the robot and fluid physics.

We first define the fluid and (single) rigid-body Lagrangians:
\begin{align}
    L^f(q^f, v^f) & = \int_\Omega \tfrac{1}{2}\rho (v^f)^Tv^f - \rho g^T q^f \, dV, \label{eq:fluid_lagrangian} \\
    L^r(q^r, v^r) & = \tfrac{1}{2}(v^r)^T M^r v^r - g^T M^r q^r, \label{eq:rigid_lagrangian}
\end{align}
where $\rho$ is the fluid density; $M^r$ is the mass matrix of the robot body; $dV$ denotes the volume differential; and $g$ is gravity. In this problem, damping is introduced by the fluid viscosity:
\begin{equation}
    F(t) = \int_\Omega \mu \Delta v^f \, dV ,
\end{equation}
where $\mu$ is the dynamic viscosity of the fluid. In a manner similar to \eqref{eq:generic_action}, we write down the action for the combine problem:
\begin{small}
    \begin{equation}
    S(q^r, q^f, v^f, v^r) = \int_{t_0}^{t_f} L^{f} + L^r + F^Tq^f + \sum_{i=1}^{6}\lambda_i c_i \, dt. \label{eq:combined_fluid_robot_action}
\end{equation}
\end{small}
The FON conditions of \eqref{eq:combined_fluid_robot_optimization} can be expressed by the variations w.r.t. $q^r, v^r$ and $q^f, v^f$, which results in the coupled incompressible Navier-Stokes and manipulator equations, respectively: 
\begin{small}
    \begin{align} 
        \rho(\dot{v}^f + (v^f \cdot \nabla)v^f) & = - \nabla p + \mu \nabla^{2}v^f - \rho g \nonumber \\
        & \quad - \frac{\partial c_3}{\partial v^f}^T \dot{\lambda}_3 - \frac{\partial c_6}{\partial v^f}^T \dot{\lambda}_6 , \label{eq:navier_stokes_momentum_coupled} \\
        \nabla \cdot v^{f} & = 0, \label{eq:navier_stokes_com} \\
        \nonumber \\
        \overbrace{M^r\dot{v^r} + M^r g}^{\text{\eqref{eq:manipulator equation}}}  & = - \frac{\partial c_5}{\partial q^r}^T \lambda_5 - \frac{\partial c_6}{\partial v^r}^T \dot{\lambda}_6. \label{eq:manipulator_velocity_coupled} \\
        \dot{q}^r & = v^r,
    \end{align}
\end{small}
where $p = \dot{\lambda}_{2} \in R^m$ is the fluidic pressure, which \citet{lagrangeMecaniqueAnalytique1853} originally realized as the dual variable that enforces the conservation of mass. \eqref{eq:navier_stokes_momentum_coupled} also provides the same constraint-based treatment of fluid-boundary conditions of \citet{allmarasLagrangeMultiplierImplementation2005}, in which coupling forces arise from the no-slip constraint at the fluid-robot interface.

\subsection{Coupled Discretization of the Fluid-Robot Multiphysics}

Using the unified least-action principle of \eqref{eq:combined_fluid_robot_optimization}, we develop a variational integrator to simulate fluid-robot multiphysics. However, we again note that $q^r$ and $v^f$ have a half-time-step lag in traditional variational integrators. Therefore, we aim to derive a new variational integrator that can integrate $q_k$ and $v_k$ while handling this temporal inconsistency.

First, we discretize the fluid domain spatially using a second-order finite-volume method~\cite{ferzigerComputationalMethodsFluid2019}, which provides a discrete counterpart to the following continuous operators:
\begin{alignat*}{2}
    \int_\Omega \rho v^f \, dV &\Rightarrow M^f v^f, \quad & \int_\Omega \nabla p \, dV &\Rightarrow Gp, \\
    \int_\Omega (v^f \cdot \nabla)v^f \, dV &\Rightarrow N(v^f), \quad & \int_\Omega \Delta v^f \, dV &\Rightarrow Lv^f.
\end{alignat*}
Through a slight abuse of notation, we will use $q^f$, $v^f$, and $p$ to refer to their discrete versions throughout the rest of this section. We may now express the spatially discrete fluid Lagrangian as:
\begin{equation}
    L^f(q^f, v^f) = \tfrac{1}{2}(v^f)^T M^f v^f - M^f g^T q^f \, dV, \label{eq:fluid_lagrangian}
\end{equation}
To handle temporal lag, we will define the constraints corresponding to $q_k$ and $\bar{v}_k$ at their respective time steps, resulting in the following discretized action:
\begin{small}
    \begin{mini}|l|
      {q^r_k, q^f_k}{\sum_{k=1}^{2} L^f_{d}(q^f_k, q^f_{k+1}) + L^r_{d}(q^r_{k}, q^r_{k+1}) + hF(t_{k} + \tfrac{h}{2})^T\bar{q}^f_{k+1}}{}{} \label{eq:discrete_fluid_robot_action}
      \addConstraint{c_2(\bar{v}^f_k) = G^T \bar{v}^f_k = 0}
      \addConstraint{c_3(\bar{v}^f_k) = B\bar{v}^f_k - v_{bc} = 0}
      \addConstraint{c_5(q^r_k) = 0}
      \addConstraint{c_6(\bar{v}^f_k, \bar{v}^r_k) = E\bar{v}^f_k - \prescript{b}{}{}\bar{v}^r(\bar{v}^r_k) = 0 ,}
    \end{mini}
\end{small}
where $B\bar{v}^f_k$ extracts fluid velocities located at the fluid-domain boundary, and $c_{6}(\bar{v}^f, \bar{v}^r)$ is expressed using the discretized immersed boundary method posed in \eqref{eq:discrete_immersed_boundary}. The external force is additionally modified to account for the convective term, $N(v^f)$:
\begin{equation}
    F(t_{k} + \tfrac{h}{2}) = L\bar{v}^f_{k+1} + N(\bar{v}^f_{k+1}).
\end{equation}
Upon inspection, we note that the derivatives of \eqref{eq:discrete_euler_lagrange} can be expressed in terms of $\Bar{q}^f$ and $\Bar{v}^r$ using the chain rule:
\begin{align}
    D_{2}L_{d}(q_{k-1}, q_{k}) & = h(\tfrac{\partial L}{\partial \Bar{q}_{k}}\underbrace{\tfrac{\partial \Bar{q}_{k}}{\partial q_k}}_{\frac{1}{2}} + \tfrac{\partial L}{\partial \Bar{v}_{k}}\underbrace{\tfrac{\partial \Bar{v}_{k}}{\partial q_k}}_{\frac{1}{h}}), \label{eq:midpoint_slot_derivative_k} \\
    D_{1}L_{d}(q_{k}, q_{k+1}) & = h(\tfrac{\partial L}{\partial \Bar{q}_{k+1}}\underbrace{\tfrac{\partial \Bar{q}_{k+1}}{\partial q_k}}_{\frac{1}{2}} + \tfrac{\partial L}{\partial \Bar{v}_{k+1}}\underbrace{\tfrac{\partial \Bar{v}_{k+1}}{\partial q_k}}_{-\frac{1}{h}}). \label{eq:midpoint_slot_derivative_k+1}
\end{align}
We can now substitute \eqref{eq:midpoint_slot_derivative_k}-\eqref{eq:midpoint_slot_derivative_k+1} into \eqref{eq:discrete_euler_lagrange}. This results in the following FON conditions for \eqref{eq:discrete_fluid_robot_action} corresponding to the robot dynamics:
\begin{small}
    \begin{align}
        q^r_{k+1} & = q^r_{k} + h\bar{v}^r_{k+1}, \label{eq:robot_config_variational} \\
        \nonumber \\
        \tfrac{1}{2}M^r\bar{v}^r_{k+1} & - h\big(\tfrac{\partial c_5}{\partial q^r_k}\big)^T \lambda_{5, k} \nonumber \\
        & - h\big(\tfrac{\partial c_6}{\partial \bar{v}^r_k}\big)^T \lambda_{6,k} = \tfrac{1}{2}M^r\bar{v}^r_k - hM^r g, \label{eq:robot_velocity_variational}
    \end{align}
\end{small}
as well as that corresponding to the fluid dynamics:
\begin{small}
    \begin{align}    
        \tfrac{1}{2}\big(M^f - \mu & hL\big)\bar{v}^f_{k+1} - \tfrac{h}{2}N(\bar{v}^f_{k+1}) \nonumber \\
        & + Gp_k + B \lambda_{3,k} - h(\tfrac{\partial c_6}{\partial \bar{v}^f_k})^T \lambda_{6,k} \nonumber \\
        & = \tfrac{1}{2}\big(M^f + \mu hL\big)\bar{v}^f_k + \tfrac{h}{2}N(\bar{v}^f_k) - hM^fg. \label{eq:fluid_velocity_variational} 
    \end{align}
\end{small}
\noindent Finally, the constraints of \eqref{eq:discrete_fluid_robot_action} form the rest of the FON conditions:
\begin{align}
    G^T \bar{v}^f_{k+1} & = 0, \label{eq:com_constraint_variational} \\
    B\bar{v}^f_{k+1} - v_{bc} & = 0, \label{eq:bc_constraint_variational} \\
    E\bar{v}^f_{k+1} - \prescript{b}{}{}\bar{v}^r(\bar{v}^r_{k+1}) & = 0, \label{eq:no_slip_constraint_variational} \\
    c_5(q^r_{k+1}) & = 0. \label{eq:robot_constraints_variational}
\end{align}
The system of equations defined by \eqref{eq:robot_config_variational}-\eqref{eq:robot_constraints_variational} can be solved using Newton's method. Specifically, $q^r_{k+1}, \bar{v}^r_{k+1}, \bar{v}^f_{k+1}$ and the corresponding dual variables can be solved for given $q^r_k, \bar{v}^r_k, \bar{v}^f_k$. Derivatives can then be computed using the implicit function theorem on the entire multiphysics problem~\cite{leeAquariumFullyDifferentiable2023}. Upon further inspection, we also note that this integration scheme results in a 2nd-order implicit leap-frog method~\cite{hairer1996solving}.
    
\subsection{An Integral-Form Immersed Boundary Method}

As previously mentioned, the immersed boundary method can be both ill-posed and ill-conditioned due to enforcing the no-slip constraint at a set of (possibly changing) discrete node locations. Therefore, we propose a new formulation that integrates the original immersed boundary method along the solid-body boundary as shown in Fig.~\ref{fig:immersed_boundary}:
\begin{equation}
    \oint_{\partial \Gamma} \int_{\Omega} v^{f} \delta (q^f - \prescript{b}{}{q}^r(q^r)) \, dVdS = \oint_{\partial \Gamma} \prescript{b}{}{v}^r(q^r, v^r) dS, \label{eq:continuous_weak_form_immersed_boundary}
\end{equation}
where $\oint dS$ denotes a surface integral. 
We discretize \eqref{eq:continuous_weak_form_immersed_boundary} with piecewise-linear interpolation along the boundary. The discrete delta function is then integrated along each linear interpolant, resulting in another convolution matrix for the following discretized immersed-boundary method:
\begin{equation}
    \Bar{E}v^f - \prescript{b}{}{v}^r(v^r) = 0, \label{eq:discrete_integral_immersed_boundary}
\end{equation}
where $\bar{E}$ is the integrated convolution matrix. Rather than explicitly formulating a forcing term such as \eqref{eq:continuous_immersed_boundary_force}, we simply differentiate through \eqref{eq:discrete_integral_immersed_boundary} as seen in \eqref{eq:robot_velocity_variational}-\eqref{eq:fluid_velocity_variational}.

\subsection{Optimization With Fluid-Robot-Multiphysics Gradients}

By leveraging its differentiability, we integrate the unified fluid-robot multiphysics into a gradient-based optimization process. We begin with the general problem:
\begin{small}
    \begin{mini}|l|
      {\beta}{J(x_{1:N})}{}{} \label{eq:combined_fluid_robot_optimization}
      \addConstraint{f(x_{k+1}, x_k; \beta) = 0,}
    \end{mini}
\end{small}
where $\beta$ are the optimization parameters; $x_{k} = [q^r, \bar{v}^r, \bar{v}^f]_k$ is the full fluid-robot state at timestep $k$; $J(x_{1:N})$ is the objective; and $f(x_{k+1}, x_k; \beta)$ are the implicit dynamics constraints of our unified fluid-robot multiphysics described by \eqref{eq:robot_config_variational}-\eqref{eq:robot_constraints_variational}. Gradient-based optimizers such as L-BFGS~\cite{nocedalNumericalOptimization2006a} solve \eqref{eq:combined_fluid_robot_optimization} using the objective gradient, $\nabla_\beta J$, which can be computed via the chain rule:
\begin{equation}
    \nabla_\beta J = \big(\frac{\partial J}{\partial x_{N}} \frac{\partial x_{N}}{\partial \beta} + \frac{\partial J}{\partial x_{N}} \frac{\partial x_{N}}{\partial x_{N-1}} \frac{\partial x_{N-1}}{\partial \beta} + \cdots \big)^T, \label{eqn:chain_rule}
\end{equation}
where $\frac{\partial x_{k}}{\partial \beta}$ and $\frac{\partial x_{k+1}}{\partial x_k}$ are computed by applying the implicit function theorem~\cite{leeAquariumFullyDifferentiable2023} to $f(x_{k+1}, x_k; \beta)$. By differentiating through the unified multiphysics, the gradients also take into account the sensitivity between the robot and fluid states.

\section{EXPERIMENTAL RESULTS} \label{sec:RESULTS}

This section presents the results of several validation studies to evaluate our fluid-robot multiphysics approach. Specifically, we look at the locomotion problem of an eel robot swimming through an initially-still fluid environment. We first realize the simulation of a steady forward-swimming gait before optimizing a C-start escape maneuver to achieve robotic turning in a desired heading direction. We compare our simulated trajectories with that of an SPH-based baseline implemented in Genesis~\cite{genesisauthorsGenesisUniversalGenerative2024a} and real-world hardware.

\begin{figure}[t!]
    \centering
    \includegraphics[width=0.46\textwidth]{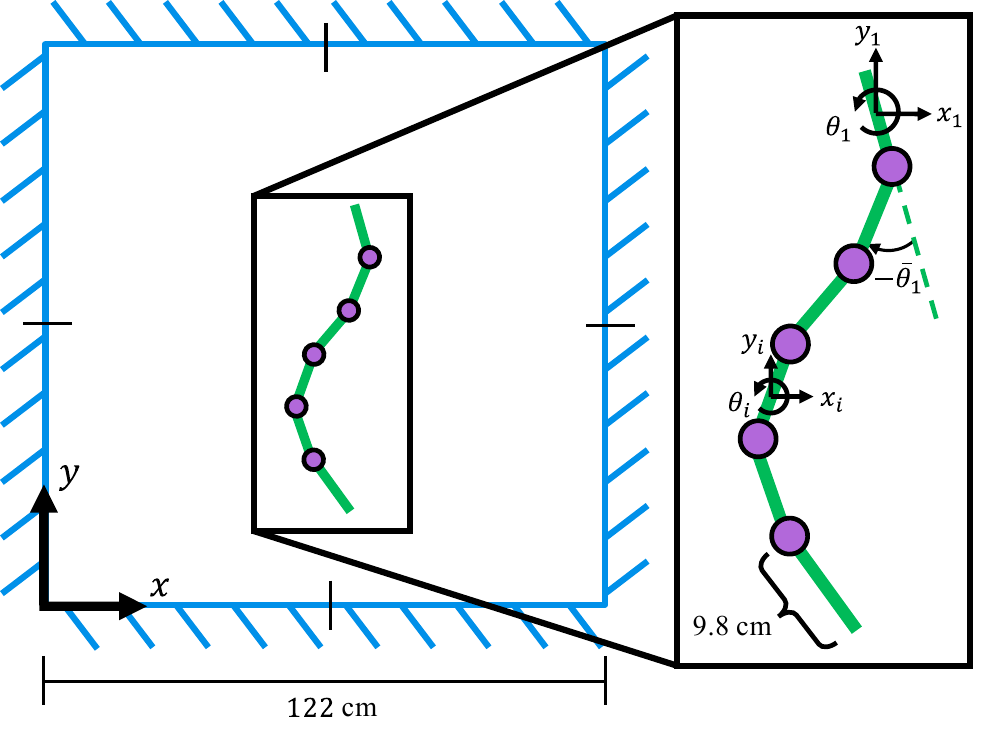}
    \caption{Our experimental setup, where a bioinspired eel robot is navigating through a walled cavity of initially still water measuring $122 \times 122$ cm. In this work, we parameterize the robot configuration in maximum coordinates where all link configurations, $[x_{i}, y_{i}, \theta_{i}]$, are specified. Joint (i.e., motor) angles between links $i$ and $i+1$ are denoted as $\bar{\theta}_{i}$.}
    \label{fig:exp_setup}
    \vspace{-\baselineskip}
\end{figure}
\begin{figure}[t!]
    \centering
    \includegraphics[width=0.48\textwidth]{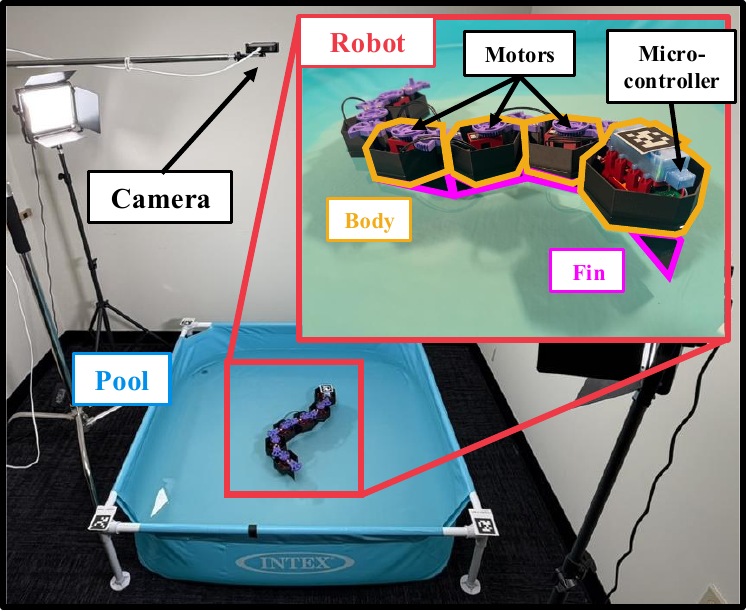}
    \caption{Experiment setup for sim-to-real validation of bioinspired swimming. The walled cavity is recreated in a pool of equal dimensions, while the eel robot is fully realized on real-world hardware. The robot hardware is fully untethered with an onboard battery, a Robotis OpenRB-150 microcontroller, and Dynamixel XC-330-M288-T servo motors to actuate the joints. 3D-printed hollow bodies keeps the eel robot afloat while thin fins are attached beneath to maintain a thin-plate profile. The head-link configuration is tracked by a top-down-facing camera via AprilTags~\cite{olsonAprilTagRobustFlexible2011a}.}
    \label{fig:hardware_setup}
    \vspace{-\baselineskip}
\end{figure}

\subsection{Experimental Setup}

As shown in Fig.~\ref{fig:exp_setup}, we simulate a swimming pool of $122 \times 122$ cm by defining a cavity of initially still water with walled boundary conditions. We utilize a 1 cm fluid cell with an equivalently-sized particle for the SPH baseline to maintain consistent fidelity. The eel robot consists of 6 segments, each 9.8 cm in length, approximated as rigid, thin plates connected at their ends via revolute joints. Servo motors drive the joints, whose PD controllers we simulate to apply joint torques. We parameterize the robot configuration in maximal coordinates, which includes all link configurations, $\{[x_{i}, y_{i}, \theta_{i}]\}^{6}_{i=1}$. We refer to joint angles as $\bar{\theta}_{i}, i=1, 2, \dots, 5$. 

We recreate the simulated environment on real-world hardware as shown in  Fig.~\ref{fig:hardware_setup}. The eel-robot hardware is designed to be fully untethered with an onboard 900-mAh battery and a Robotis OpenRB-150 microcontroller on the head (i.e., first) link. Subsequent links consist of Dynamixel XC330-M288-T servo motors to drive the joint angles. The electronics hardware are housed in buoyant, 3D-printed bodies to avoid contact with the fluid. Thin, rigid fins are attached underneath the bodies to replicate the thin-plate profile realized in simulation. A camera is used to track the configuration of the head link using AprilTags~\cite{olsonAprilTagRobustFlexible2011a}.

\begin{figure*}[t!]
    \centering
    \begin{subfigure}[t]{0.31\textwidth}
        \includegraphics[width=\textwidth, height=5.5cm]{figures/figure8a_x_error_comparison}
        \label{fig:p_flow_setup}
    \end{subfigure}
    \quad
    \begin{subfigure}[t]{0.31\textwidth}
        \includegraphics[width=\textwidth, height=5.5cm]{figures/figure8b_y_error_comparison}
        \label{fig:disc_setup}
    \end{subfigure}
    \quad
    \begin{subfigure}[t]{0.31\textwidth}
        \includegraphics[width=\textwidth, height=5.5cm]{figures/figure8c_theta_error_comparison}
        \label{fig:squid_setup}
    \end{subfigure}
    \vspace{-\baselineskip}
    \caption{Average root-mean-square error (RMSE) and associated standard deviations for simulated versus hardware-validated trajectories of an eel robot's head-link configuration during steady undulatory swimming. This is performed for joint-angle amplitudes of $10^{\circ}$, $20^{\circ}$, $30^{\circ}$, and $40^{\circ}$ across 20 trials per amplitude. Our unified fluid-robot multiphysics (\color{jj_green}green\color{black}) consistently outperforms the smoothed-particle-hydrodynamics baseline (\color{jj_orange}orange\color{black}) across all undulation amplitudes, with the improvement becoming more pronounced at higher amplitudes. At 40° amplitude, we achieve error reductions as high as 75\% compared to the baseline, demonstrating that strongly coupled fluid-robot simulation significantly improves fidelity for robotic swimming.}
    \label{fig:forward_swimming_rmse}
\end{figure*}
\begin{figure*}[h!]
    \centering
    \includegraphics[width=0.99\textwidth]{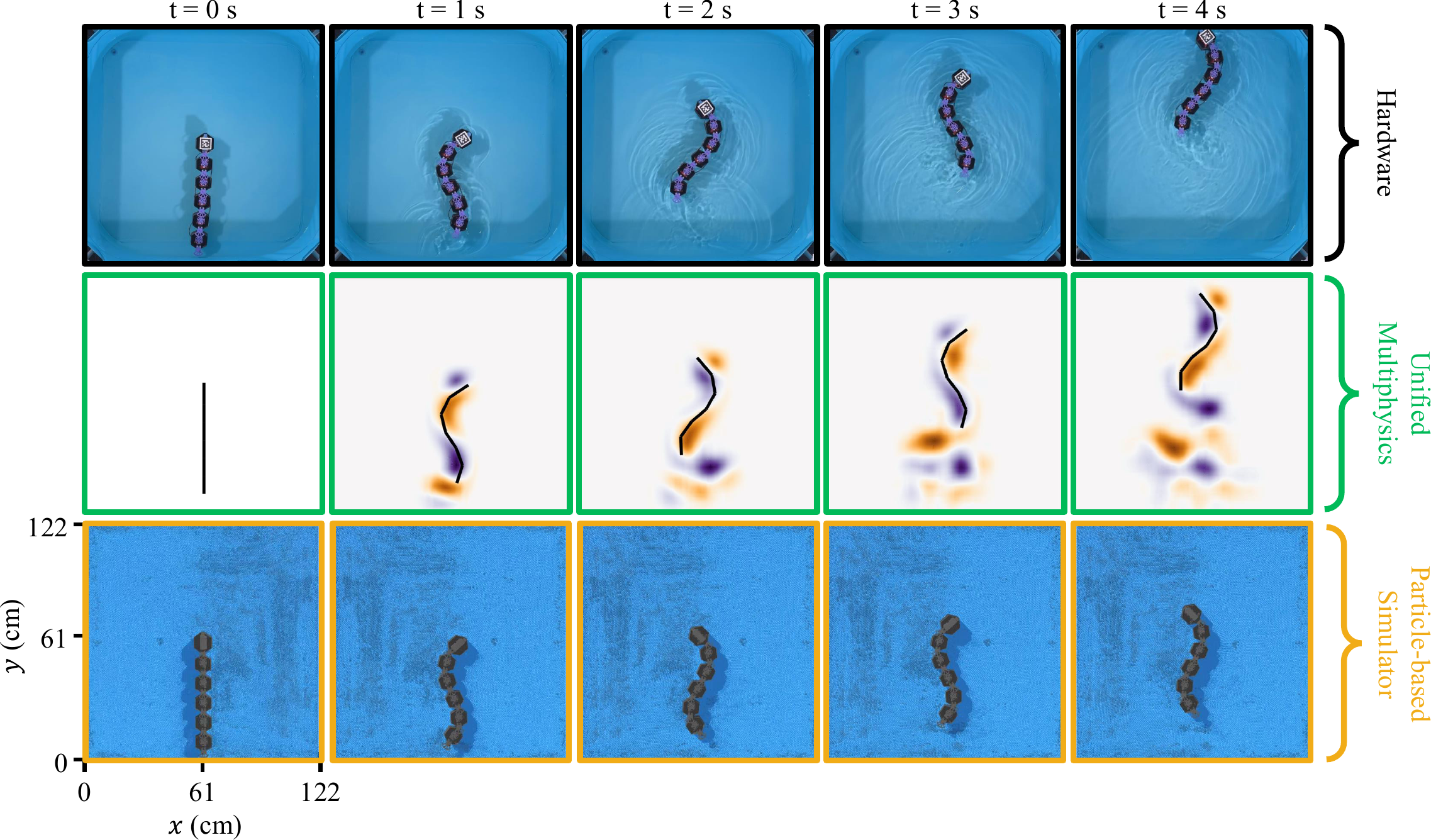}
    \caption{Time-lapse sequences showing a multilink eel-robot executing an undulating gait with a $40^{\circ}$ joint-angle amplitude. The top row (black) demonstrates the real-world robot, while the middle row (\color{jj_green}green\color{black}) is our method and the bottom row (\color{jj_orange}orange\color{black}) is the smoothed-particle-hydrodynamics-based (SPH-based) baseline. While our cell-based method is able to closely follow the forward-swimming seen in the real world, the SPH-based baseline achieves comparatively little forward progress.}
    \label{fig:forward_swimming_visualization}
    \vspace{-\baselineskip}
\end{figure*}

\subsection{Steady Undulatory Swimming}

We first perform steady undulation to realize forward swimming of the eel robot. The joint angles are driven by a sinusoidal profile with the same amplitude, each with the appropriate phases to create a traveling wave that undulates at $1$ hz with a wavelength equal to the length of the robot body. In this study, we evaluate swimming performance across multiple amplitudes of $10^\circ, 20^\circ, 30^\circ, 40^\circ$. Specifically, we compute the root-mean-square error (RMSE) across a 4-s trajectory of the head link configuration. This was performed over 20 trials for each amplitude to compute the mean RMSE and standard deviations reported in Fig.~\ref{fig:forward_swimming_rmse}.

As shown in Fig.~\ref{fig:forward_swimming_rmse}, our unified-multiphysics approach is able to qualitatively realize the forward-swimming motion seen on hardware for a $40^\circ$ amplitude. Meanwhile the SPH baseline achieves comparatively little forward progress. This appears to be the case for all amplitudes, with our method consistently achieving much lower RMSE error. In the forward-swimming direction ($y$), the baseline's RMSE error appears to increase with amplitude while our method maintains a more consistent, slightly decreasing, trend. This results in a nearly $75\%$ improvement for the $40^\circ$ case and suggests that the SPH baseline is unsuitable for more dynamic swimming. We attribute this to the particle-based fluid representation's poor satisfaction of physical constraints such as incompressibility and volume conservation~\cite{suchdeVolumeMassConservation2025, nairVolumeConservationIssues2015}.

\begin{figure}[t!]
    \centering
    \includegraphics[width=0.48\textwidth]{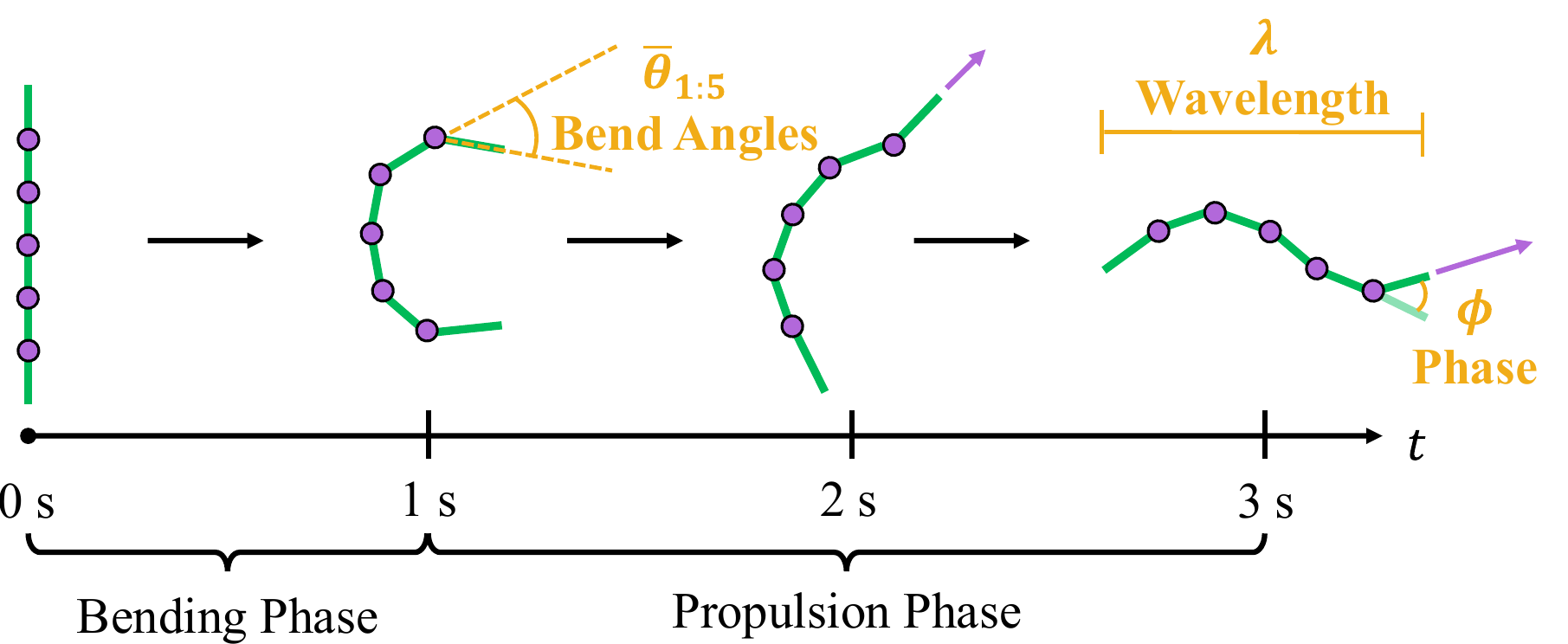}
    \caption{Kinematic sequence of the eel robot performing a C-start escape maneuver to achieve rapid turning. During the bending phase (0–1 second), the robot curls its body into a characteristic C-shape parameterized by the bending angles at the joints, $\theta_{1:5}$. In the propulsion phase (1–3 seconds), the body rapidly straightens and propagates a traveling wave toward the tail, generating a strong fluid interaction to accelerate in a new heading direction (\color{jj_purple} purple arrows\color{black}). This is parameterized by the the traveling wave's phase, $\phi$ and wavelength, $\lambda$. In this work, we aim to optimize this gait (\color{jj_orange}orange\color{black}) to perform a 90-degree turn from initial rest.}
    \label{fig:c_start_setup}
    \vspace{-\baselineskip}
\end{figure}

\subsection{Optimizing a Highly Dynamic C-Start Maneuver}

We leverage our method's differentiability to optimize a highly dynamic C-start escape maneuver inspired by fish~\cite{gazzolaCstartOptimalStart2012, schusterArcherfishPredictiveCstart2023, domeniciEscapeResponsesFish2019}. As shown in Fig.~\ref{fig:c_start_setup}, the maneuver involves the robot curling into a C-shape from an initial vertical orientation during a bending phase (0-1 seconds), followed by a powerful transition into undulation during the propulsion phase (1-3 seconds) to achieve fast reorientation and acceleration. We optimize the gait to perform a $90^\circ$ turn, which is expressed by the following quadratic objective:
\begin{equation}
    J(x_{N}) = \frac{1}{2}(q^\text{com}_{N} - q^\text{goal})^TQ(q^\text{com}_{N} - q^\text{goal}), \label{eq:c_start_objective}
\end{equation}
where $q^\text{com}_{N}$ is the robot's final-time center-of-mass configuration; $q_{\text{goal}}$ is the goal configuration representing a successful $90^\circ$ turn; and $Q$ is the cost matrix. We optimize over the joint bending angles, $\bar{\theta}_{1:5}$ during the bend phase, the initial phase of $\bar{\theta}_1$, $\phi$, and the wavelength of the undulating wave, $\lambda$, We fix the amplitude per joint to $25^\circ$ to prevent the robot from hitting the wall. The optimization is performed using the gradient-based box-constrained L-BFGS algorithm~\cite{nocedalNumericalOptimization2006a} with bounds of $10^\circ \leq \bar{\theta}_{1:5} \leq 36^\circ$, $-\frac{\pi}{2} \leq \phi \leq 0$ rad, and $0.5L^B \leq \lambda \leq L^B$, where $L^B$ is the body length of the robot. We begin the optimization with an initial guess of $\bar{\theta}_{1:5}^\text{ig} = 10^\circ$, $\phi^\text{ig} = -\tfrac{\pi}{3}$ rad, and $\lambda^\text{ig} = 0.6L^B$. The optimized values are $\bar{\theta}_{1:5}^* = [17^\circ, 18^\circ, 20^\circ, 24^\circ, 36^\circ]$, $\phi^* = -1.16$ rad, and $\lambda^* = 0.57L^B$. The resulting gaits are rolled out in an open-loop manner, and the corresponding time-lapse sequences are visualized in Fig.~\ref{fig:c_start_trajectory_visual}. 

As shown in Fig.~\ref{fig:c_start_trajectory_visual}, our optimization process is able to calculate gait parameters that successfully realize the desired $90^\circ$ turn. We do note that our method consistently undershoots the travel distance observed in the real world, which is likely due to unmodeled free-surface dynamics, servo compliance, thin-plate geometry approximations, and uncertainty in physical parameters such as robot mass properties and fluid viscosity. Despite this sim-to-real gap, the $90^\circ$ turn is still fully realized on hardware. Meanwhile, the SPH baseline fails to realize both the appropriate turning and propulsion, further suggesting its unsuitability for highly dynamic robot-swimming tasks.


\subsection{Computational Cost}

Regarding computational cost, a 3-second C-start simulation at a 100 Hz sample rate in our unified multiphysics framework requires approximately 261 seconds (0.87 seconds per timestep) on the CPU of an M3-Ultra Mac Studio (28 cores). The same rollout in Genesis~\cite{genesisauthorsGenesisUniversalGenerative2024a} requires approximately 2727 seconds (9.09 seconds per timestep) on the M3-Ultra CPU and 300 seconds (1 seconds per timestep) on an NVIDIA RTX 3070-Ti GPU.

\section{LIMITATIONS} \label{sec:LIMITATIONS}

Our unified framework for modeling strongly coupled multiphysics has several limitations: First, the computational complexity of performing implicit time integration is high, especially when extended to 3D. Future work is needed to develop efficient sparsity-exploiting solvers to scale our method to large-scale problems. Additionally, while the Eulerian fluid representation is well suited for dynamic tasks like swimming, it is less convenient for visual rendering than particle-based methods. 

\section{CONCLUSIONS} \label{sec:CONCLUSION}

We have presented a unified framework for deriving and simulating fluid-robot multiphysics as a single optimization problem. Specifically, we model the unified least-action principle from which the coupled differential equations are derived. We build upon previous works in variational mechanics to discretize the action directly to simulate the fluid-robot interaction in a stable, tightly coupled manner. We additionally modify the original immersed-boundary method to handle articulated multibody systems found in robotics. We validate our approach on a bioinspired eel robot, where we realize both undulatory swimming and fast C-start escape maneuvers via gradient-based optimization. Compared to a weakly coupled, particle-based baseline, the resulting swimming behaviors align much more closely with real-world experiments, showcasing the importance of representation and coupling for improved sim-to-real transfer in fluid-robot multiphysics settings.

In future work, we aim to address the computational scalability of our variational method for downstream design and trajectory-optimization tasks. We also wish to explore the connections to other CFD methods like implicit large-eddy simulations. Finally, we hope to extend our framework to other multiphysics problems such as rigid-soft robots that experience contact.



\balance
\bibliographystyle{plainnat}
\bibliography{variational_fsi}

\end{document}